\documentclass{esannV2}
\usepackage{graphicx}
\usepackage[latin1]{inputenc}
\usepackage{amssymb,amsmath,array}

\usepackage{pifont}
\usepackage{dsfont}

\usepackage{amsfonts}
\usepackage{amssymb}
\usepackage{amsthm}

\usepackage{float}
% ...
\inputencoding{latin1}
% ...
\inputencoding{utf8}
\usepackage{mwe}
\usepackage{caption}

\newtheorem{proposition}{Proposition}
\usepackage{etoolbox}

\begin{document}
\title{Equilibrium Propagation for \\ Complete Directed Neural Networks}
\author{Matilde Tristany Farinha$,^{1,3}$ Sérgio Pequito$,^2$\\
Pedro A. Santos$,^1$ Mário A. T. Figueiredo$^3$
\vspace{.3cm}
\\
$1$ - INESC-ID \& Dept. of Mathematics, IST, University of Lisbon, Portugal
\vspace{.1cm}\\
$2$ - Dept. of Industrial and Systems Engineering, RPI, Troy (NY), USA
\vspace{.1cm}\\
$3$ - Instituto de Telecomunicações, IST, University of Lisbon, Portugal
}

\maketitle
\footnotetext[1]{This work was supported by national funds through FCT, Fundação para a Ciência e a Tecnologia, under projects UIDB/50021/2020 and SLICE PTDC/CCI-COM/30787/2017, and the first author has conducted part of this research as a visiting research scholar in the Department of Industrial Systems and Engineering, Rensselear Polytechnique Institute.}
\addtocounter{footnote}{1}

\begin{abstract}
    Artificial neural networks, one of the most successful approaches to supervised learning, were originally inspired by their biological counterparts. However, the most successful learning algorithm for artificial neural networks, backpropagation, is considered biologically implausible. We contribute to the topic of biologically plausible neuronal learning by building upon and extending the equilibrium propagation learning framework. Specifically, we introduce: a new neuronal dynamics and learning rule for arbitrary network architectures; a sparsity-inducing method able to prune irrelevant connections; a dynamical-systems characterization of the models, using Lyapunov theory.
\end{abstract}

\section{Introduction} \label{sec:introduction}
    \vspace{-0.1cm}
    Nowadays, many state-of-the-art approaches to \textit{supervised learning} rely on \textit{artificial neural networks} (ANNs). \textit{Backpropagation} (BP) \cite{Rumelhart1987LearningPropagation}, the most successful algorithm to train ANNs \cite{Lecun2015DeepLearning}, is considered bio-implausible since: (\emph{i}) it lacks local error representation; (\emph{ii}) it uses distinct forward and backward information passes; (\emph{iii}) it requires symmetric synaptic weights \cite{Guerguiev2017TowardsDendrites}. To bridge the gap between biological and machine learning, it is thus crucial to find alternatives to BP-based algorithms that encompass properties of biological neural networks, a goal to which much research effort has been devoted.\par
    
    The \textit{equilibrium propagation} (EP \cite{Scellier2017EquilibriumBackpropagation}) model adopts a local learning rule and uses just one kind of computation both for the forward and backward information passes, therefore being a more bio-plausible alternative to BP. However, EP has some bio-unrealistic aspects: (\emph{i}) it assumes symmetric synaptic weights; (\emph{ii}) it has only been tested on layered architectures; and (\emph{iii}) it does not promote sparse networks \cite{OReilly1998SixCognition}. We tackle these problems by introducing the \textit{DirEcted EP} (DEEP) learning framework, which: (\emph{i}) allows for asymmetric feedback weights; (\emph{ii}) allows for arbitrary network architectures; and (\emph{iii}) promotes sparse networks. DEEP assumes the network is an arbitrary complete directed graph and its training algorithm actively removes the presumably expendable connections by inducing sparsity. We also establish sufficient conditions for convergence of the neuronal dynamics of DEEP's inference phase.\par
    
    The remainder of this paper is organized as follows: in Section 2, state-of-the-art EP-like models are discussed; in Section 3, the new DEEP model is introduced; in Section 4, DEEP is experimentally evaluated and the results are analyzed; and in Section 5, conclusions and possible future work are discussed.

\section{Related Work} \label{sec:related_work}
    \vspace{-0.1cm}
    DEEP was inspired by the original EP, which is an energy-based model described as a multi-layered continuous Hopfield network of recurrently connected neurons with symmetric weights \cite{Scellier2017EquilibriumBackpropagation}. The idea is to mimic a central aspect of the behaviour of the brain, where neurons perform local computations and evolve collectively towards an equilibrium state of the corresponding dynamical system.\par
    
    Several extensions and adaptations of the original EP have been proposed. For instance, an asymmetric version of EP (with asymmetric synaptic weights) was proposed \cite{Scellier2018GeneralizationDynamics}; however, with a complete graph architecture, we observed experimentally that the asymmetric EP model is sometimes unable to learn. A bidirectional-EP working both as a generative and a discriminative model was also proposed; however, although it provides an insightful extension of EP \cite{Khan2018BidirectionalWaterloo}, it does not solve the weight-transport problem \cite{Grossberg1987CompetitiveResonance}. Another EP adaptation considers spiking neurons, therefore taking an important step towards bio-plausibility \cite{OConnor2019TrainingPropagation}. Additionally, EP has been extended to convolutional architectures, for which the lowest error rate among EP-like models on the MNIST classification task \cite{LeCun2010MNISTDatabase} has been reported -- ``approximately $1\%$" \cite{Ernoult2019UpdatesInput}. However, this did not improve the model's bio-plausibility, since convolutional architectures are considered bio-implausible due to their extensive weight sharing \cite{Bartunov2018AssessingArchitectures}.

\section{DirEcted Equilibrium Propagation (DEEP)} \label{sec:model}
    \vspace{-0.1cm}
    The DEEP model, for an architecture with a total of $N$ neurons (from which $P$ are input neurons), is completely specified by the following elements:
    \begin{itemize}\itemsep0.07cm
    \vspace{-0.1cm}
    \item a state vector $\boldsymbol{s}(t)=[s_{j}(t)]_{j=1}^{N}\in[0,1]^{N}$, containing the neuronal activities, \textit{i.e.}, ``firing rates"; this state vector includes sub-vectors that correspond to (fixed) input $\boldsymbol{x} = [s_{j}(t)]_{j=1}^{P}$, hidden $\boldsymbol{h}(t)$, and output $\hat{\boldsymbol{y}}(t)$ neurons;
    \item a matrix $\boldsymbol{W}=[W_{ij}]_{i,j=1}^{N}$, where $W_{ij}$ is the weight associated to the connection from neuron $i$ to $j$;
    \item a bias vector $\boldsymbol{b}=[b_{j}]_{j=1}^{N}$, where $b_{j}=0$ for every $j:s_j(t)\in\boldsymbol{x}$; and
    \item a set of continuous-time differential equations defining its dynamics.
    \end{itemize}
    
    Henceforth, the time dependency of the state variable is omitted to shorten the notation. Defining $\theta=(\boldsymbol{W},\boldsymbol{b})$ and with the input neurons fixed, the neuronal dynamics we propose is dictated by the vector fields (recall that $\dot{s}_{j}=d s_j/dt$)
    \vspace{-0.1cm}
    \begin{equation} \label{eq:neuronal_dyn_ours}
        \dot{s}_{j} = V_{\theta,j}^{\beta}(\boldsymbol{s},\boldsymbol{y},\beta) = \sum\limits_{i=1}^{N}W_{ij}s_{i} + b_{j} - s_{j}\sum\limits_{i=1}^{N}W_{ji} - \beta\, \frac{\partial C_{\theta}(\hat{\boldsymbol{y}},\boldsymbol{y})}{\partial s_{j}}\mathds{1}_{\hat{\boldsymbol{y}}}(s_{j}),
    \end{equation}
    where $\boldsymbol{y}$ is the vector of target/desired outputs, $C_{\theta}(\hat{\boldsymbol{y}},\boldsymbol{y})$ is the cost (e.g., mean squared error -- MSE), $\beta$ controls how much the cost influences the dynamics, and $\mathds{1}_{\hat{\boldsymbol{y}}}(s_{j}) = 1$ if and only if $s_j\in\hat{\boldsymbol{y}}$ (\textit{i.e.}, if $s_j$ is an output neuron), otherwise it is zero. Note that this new dynamics accounts for the weighted incoming and outgoing connections and the leakage of a neuron's activity depends not only on its own activity (as in the original and asymmetric versions of the EP \cite{Scellier2017EquilibriumBackpropagation,Scellier2018GeneralizationDynamics}), but also on the weighted sum of outgoing connections.\par
    
    The training algorithm has two distinct phases: the first phase, with $\beta=0$; and the second phase, with $\beta\neq0$. In the first and second phases, the network settles to equilibrium states, denoted $\boldsymbol{s}^{0}$ and $\boldsymbol{s}^{\beta}$, respectively \footnote{To ensure that the firing rates are bounded by $[0,1]$ and facilitate convergence, after each discrete update of the state variable $\boldsymbol{s}$, the state variable is bounded by the hard-sigmoid function.}. The loss function is defined as the cost when the network is at its first equilibrium state, \mbox{$J_{\theta}(\boldsymbol{x},\boldsymbol{y}) = C_{\theta}(\boldsymbol{s}^{0},\boldsymbol{y})$}. When used for inference, the activities of the input neurons are fixed and the network evolves to equilibrium $\boldsymbol{s}^{0}$, from which the output is read at the corresponding output neurons $\hat{\boldsymbol{y}}$.\par
    
    The learning rule we propose is obtained by numerical integration of the bio-inspired weight dynamics $\dot{W}_{ij}\propto s_{i}(t)\dot{s}_{j}(t)$ \cite{Xie1999Spike-basedActivity} in the path from $\boldsymbol{s}^{0}$ to $\boldsymbol{s}^{\beta}$ in $M_{\beta}$-steps (the time derivative $\dot{s}_{j}$ is approximated by a backward difference):
    \vspace{-0.1cm}
    \begin{equation} \label{eq:W_update_our}
        \Delta{W}_{ij} \propto \frac{1}{M_{\beta}}\sum_{m=M_{0}+1}^{M_{0}+M_{\beta}}s_{i}(m)\big(s_{j}(m)-s_{j}(m-1)\big),
    \end{equation}
    where $M_0$ is the number of steps of the first phase. Interpreting the biases as the weighted outgoing connections from a neuron with activity fixed to one, their updates are also given by Equation (\ref{eq:W_update_our}). Note that the learning rule of the previously proposed asymmetric version of EP \cite{Scellier2018GeneralizationDynamics} corresponds to a one-step version of the numerical integration in Equation (\ref{eq:W_update_our}) with a forward difference approximation of $\dot{s}_{j}$, which yields simply $\Delta W_{ij}\propto s_{i}^{0}(s_{j}^{\beta}-s_{j}^{0})$.\par
    
    Sparsity-inducing $\ell_{1}$ regularization is added to the proposed learning rule. Moreover, sparsity is actively enforced at each weight update when such weight is below a certain threshold. Specifically, each weight $|W_{ij}|<\lambda$, with $\lambda>0$ small, is randomly removed from the network with probability $p_{ij}$, given by a Boltzmann distribution defined across the incoming weights of neuron $j$, i.e.,
    \vspace{-0.1cm}
    \begin{equation} \label{eq:prob_ij}
        p_{ij} = e^{-|W_{ij}|/T}\Big/\sum\limits_{k=1}^{N}e^{-|W_{kj}|/T}.
    \end{equation}
    
    In this context, the temperature $T$ represents how likely it is for stronger connections (in absolute value) to be deemed irrelevant.
    
\subsection{Analytic Properties} \label{subsec:analytic_prop}
    \vspace{-0.1cm}
    \subsubsection{Time-invariant Sum of Firing Rates} \label{subsubsec:firing_rates_sum_inv}
        \vspace{-0.1cm}
        DEEP's neuronal dynamics is bio-plausible as it has been reported that, in the absence of external sensory stimulus or motor activity, the grand mean firing rate of the hippocampal neurons remains constant \cite{Hirase2001FiringExperience}.
        
        \vspace{-0.1cm}
        \begin{proposition} \label{prop:bio_property}
            In the absence of external stimulus (no fixed activity of the input or bias neurons), the dynamics in Equation (\ref{eq:neuronal_dyn_ours}), with $\beta=0$, preserves the sum of firing-rates through time \mbox{(\textit{i.e.}, $\sum_{j}\dot{s}_{j}(t)=0,\,t\in\mathbb{R}_{0}^{+}$)}.
        \end{proposition}
        \vspace{-0.35cm}
        \begin{proof}
            It follows from $\sum_{j=1}^{N}\dot{s}_{j} = \sum_{j=1}^{N}V_{\theta,j}^{0}(\boldsymbol{s}) = \sum_{i,j=1}^{N}(W_{ij}s_{i}-W_{ij}s_{i}) = 0$.
        \end{proof}
        
    \vspace{-0.4cm}
    \subsubsection{Sufficient Conditions for Stability} \label{subsubsec:gen_stab_test}
        \vspace{-0.1cm}
        Sufficient conditions for local asymptotic stability of $\boldsymbol{s}^{0}$ (i.e., the equilibrium state reached during inference) can be obtained by leveraging on Gergschorin's circle theorem and nonlinear control analysis tools from Lyapunov theory \cite{Bejarano2018ACircles}. Specifically, sufficient conditions for the stability of the inference phase of DEEP are given in the following proposition.
        \vspace{-0.1cm}
        \begin{proposition} \label{prop:our_stability_test}
            If $\sum_{i=1}^{N}W_{ji}>0$ and $\sum_{i=P+1}^{N}\left|W_{ij}\right|<\left|\sum_{i=1}^{N}W_{ji}\right|$ is verified for $j\in\{P+1,...,N\}$, then $\boldsymbol{s}^{0}$ (equilibrium state with respect to Equation (\ref{eq:neuronal_dyn_ours}), with $\beta=0$) is locally asymptotically stable.
        \end{proposition}
        \vspace{-0.35cm}
        \begin{proof}
        For a time-invariant nonlinear dynamical system $\dot{\boldsymbol{s}}(t)=f(\boldsymbol{s}(t))$, an equilibrium state $\boldsymbol{s}^{*}$ is locally asymptotically stable if $J_{jj}<0$ and $R_{j}<|J_{jj}|$ is satisfied for $j\in\{1,\ldots,N\}$, where $\boldsymbol{J}=Df(\boldsymbol{s}^{*})\in\mathbb{R}^{N\times N}$ (the Jacobian of $f$ evaluated at $\boldsymbol{s}^{*}$), and $R_{j}=\sum_{i=1,i\neq j}^{N}|J_{ji}|$ \cite{Bejarano2018ACircles}. These conditions can be particularized for the system given by Equation (\ref{eq:neuronal_dyn_ours}) with $\beta=0$, where $\boldsymbol{J}=[\partial V_{\theta,j}^{0}(\boldsymbol{s}(t),\boldsymbol{y},\beta)/\partial s_{j}]_{j}$, for $j\in\{P+1,...,N\}$, yielding the conditions stated in the proposition.
        \end{proof}

\section{Experiments and Results} \label{sec:experiments}
    \vspace{-0.1cm}
    DEEP is constrained by the curse of dimensionality due to its high dimensional search space, so its performance is analyzed for simple tasks such as learning logical operations. For these tasks, the architecture considered is a 8-neuron complete directed graph. Although DEEP can learn XOR with 1 hidden neuron and AND and OR with none, 5 hidden neurons are used so that sparsity is perceivable when using the sparsity-inducing method.
    
    \vspace{-0.1cm}
    \begin{figure}[H]
    \centering
    \begin{minipage}{.47\textwidth}
    \centering
    \includegraphics[width=\linewidth]{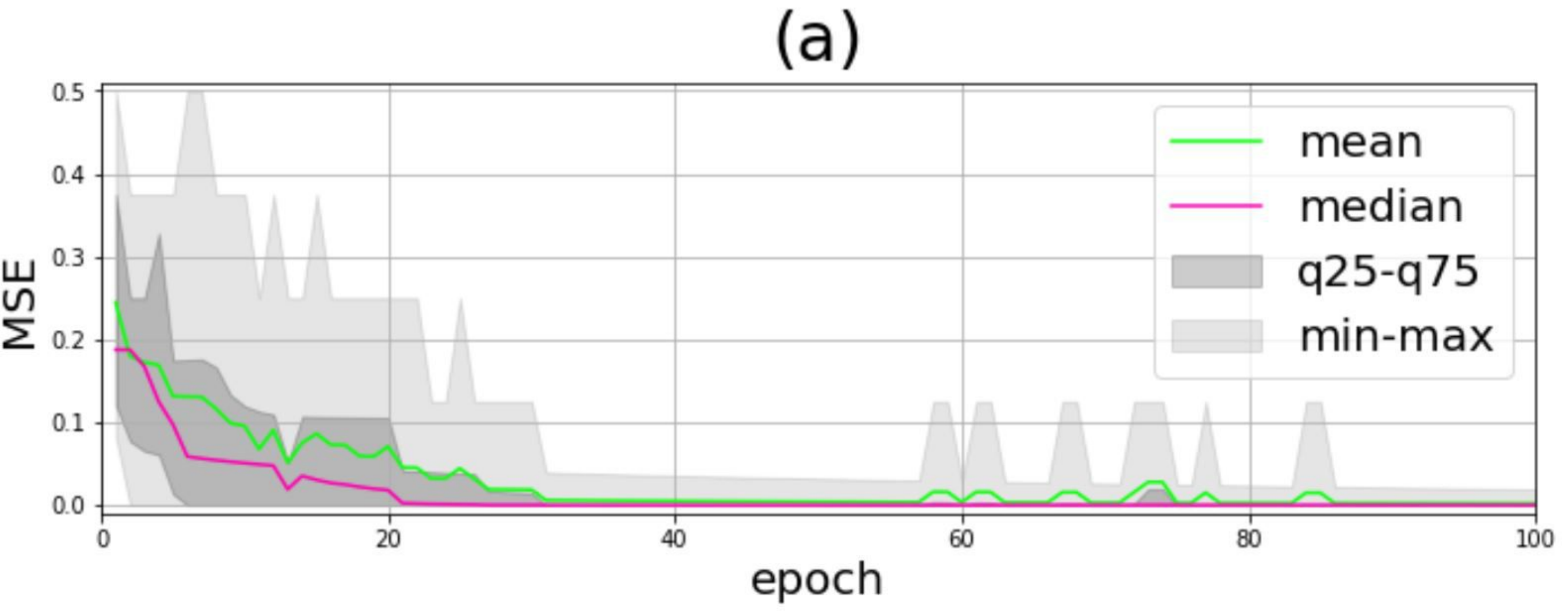}
    \end{minipage}
    \begin{minipage}{.47\textwidth}
    \centering
    \includegraphics[width=\linewidth]{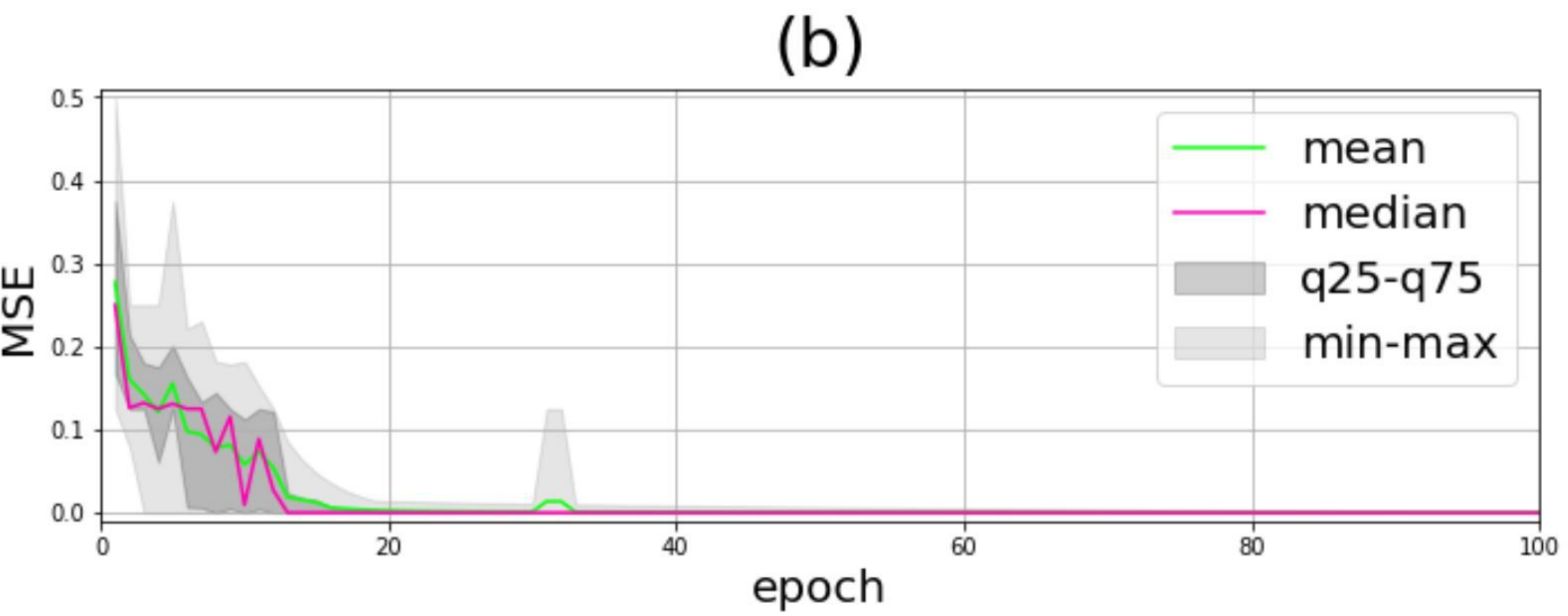}
    \end{minipage}
    \centering
    \includegraphics[width=0.96\linewidth]{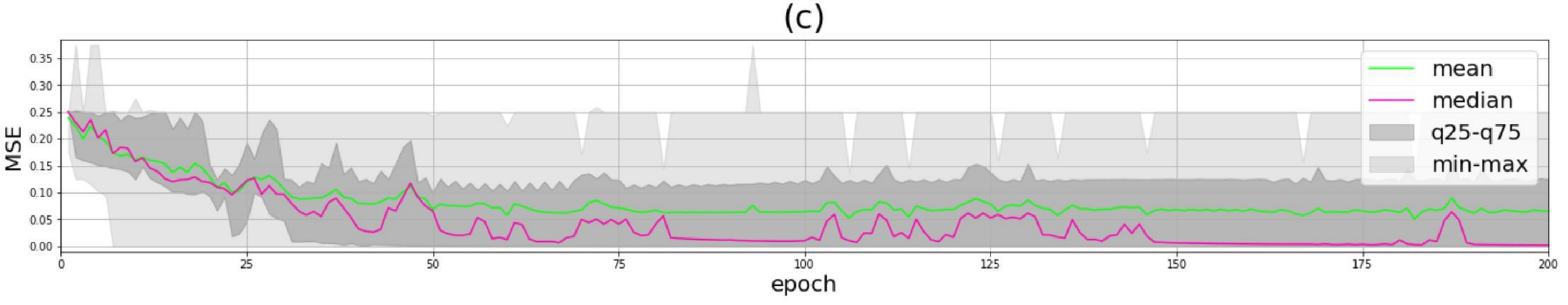}
    \caption{MSE convergence during the first phase of 10 independently trained models while learning the logical operations: (a) AND, (b) OR, and (c) XOR.}
    \label{fig:mse}
    \end{figure}
    
    \begin{figure}[H]
    \centering
    \begin{minipage}{.22\textwidth}
        \centering
        \includegraphics[width=\linewidth]{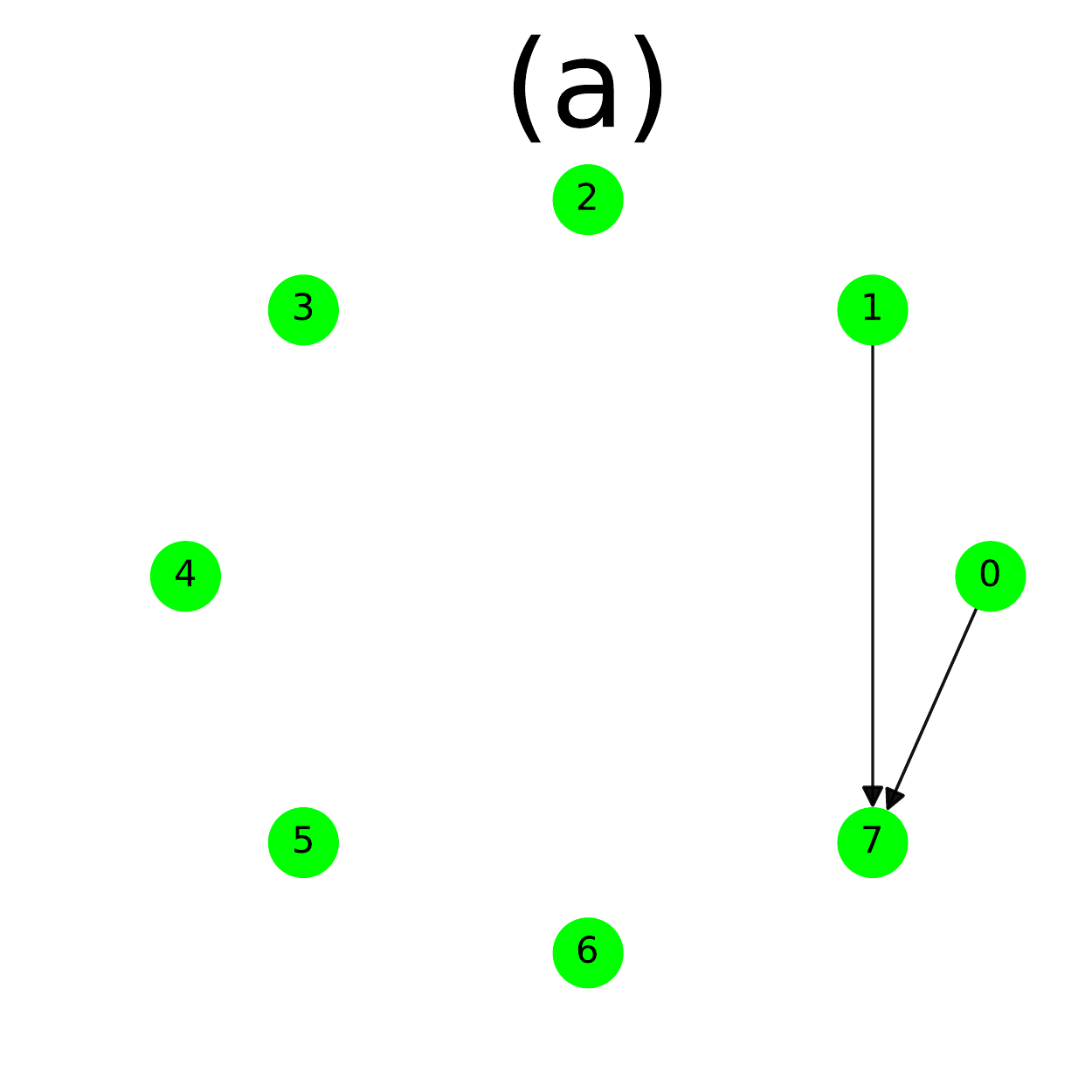}
    \end{minipage}
    \begin{minipage}{.22\textwidth}
        \centering
        \includegraphics[width=\linewidth]{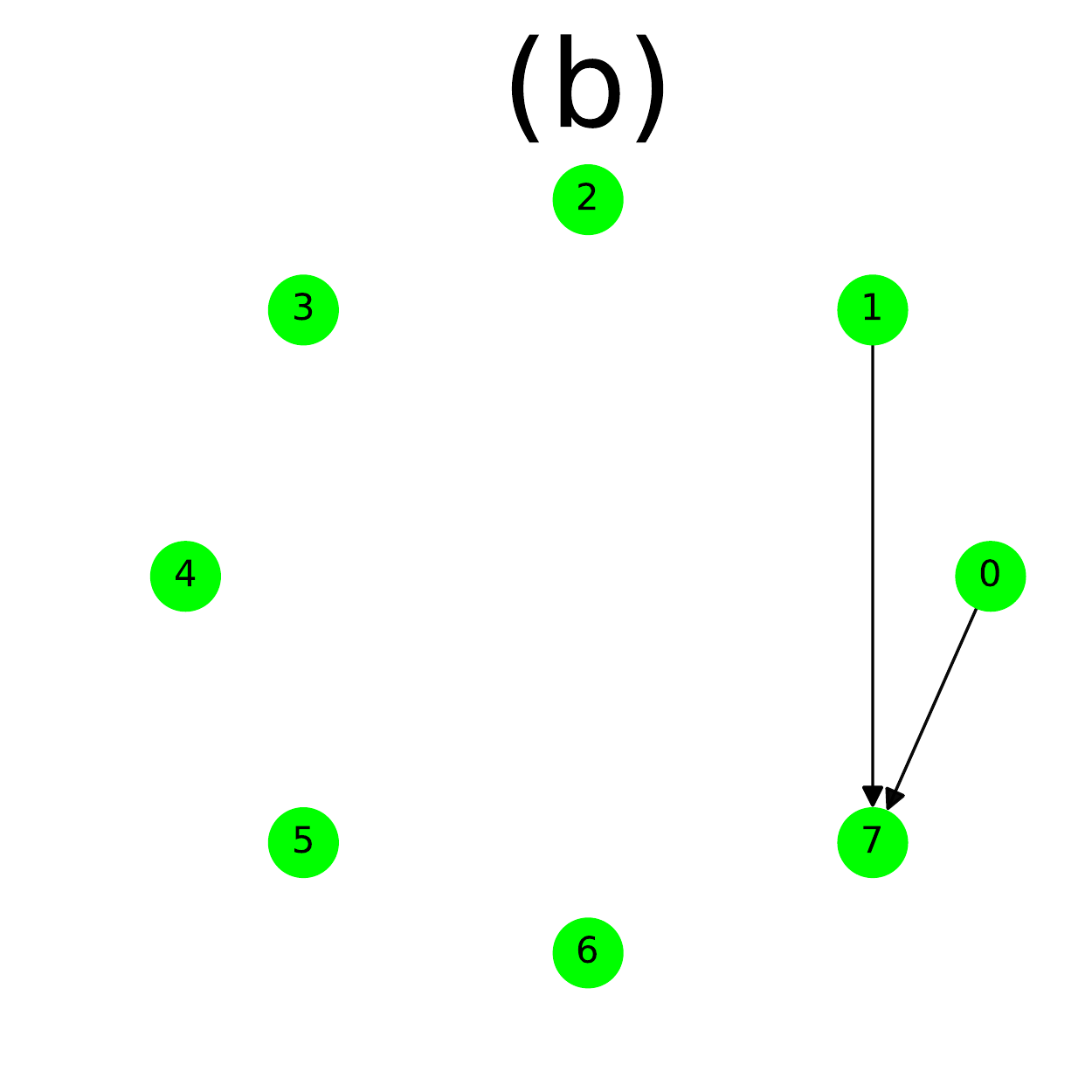}
    \end{minipage}
    \begin{minipage}{.22\textwidth}
        \centering
        \includegraphics[width=\linewidth]{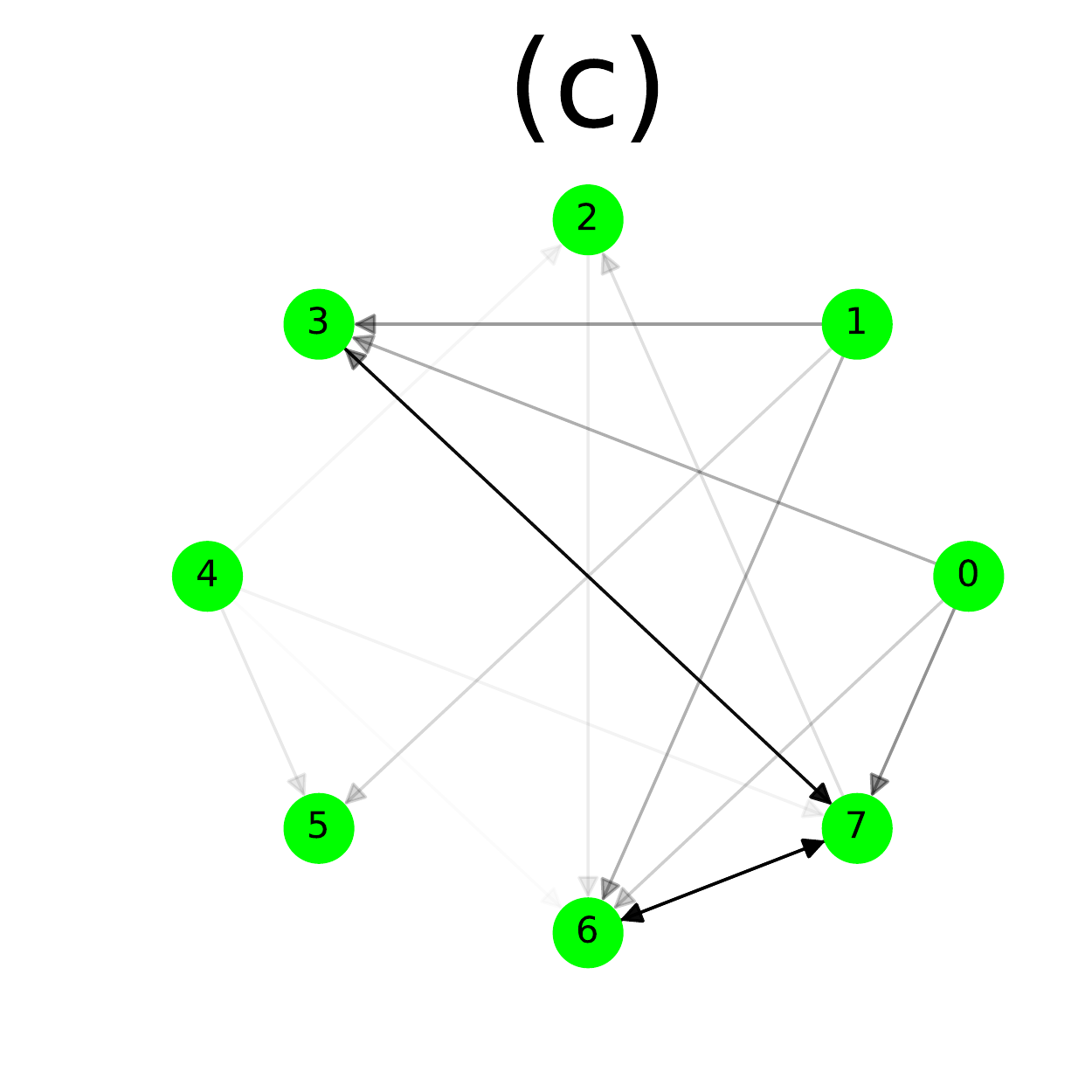}
    \end{minipage}
    \caption{Sparse networks trained for learning the logical operations: (a) AND ($93.75\%$ sparse), (b) OR ($95.83\%$ sparse), and (c) XOR ($62.50\%$ sparse). The indexes refer to the neuron's type: \textcircled{\footnotesize{0}}-\textcircled{\footnotesize{1}} input, \textcircled{\footnotesize{2}}-\textcircled{\footnotesize{6}} hidden, \mbox{\textcircled{\footnotesize{7}} output}; the connections' opacity is proportional to their strength (in absolute value).}
    \label{fig:sparse}
    \end{figure}
    
    Figure \ref{fig:mse}\footnote{The min-max refers to the interval of values contained between the minimum and the maximum of the data, and the q25-q75 refers to the interval of values contained between the first and third quartiles of the data.} illustrates how, when learning XOR, the MSE converges to zero most of the times but not always. Besides, it converges less frequently than when learning AND or OR, which highlights how XOR is harder to learn. Additionally, DEEP is also trained with the sparsity-inducing method mentioned above. We observed that, with this method, the most relevant connections are strengthened and, while for simple tasks all the expendable connections are removed, for more complex tasks only a few are removed (see Figure \ref{fig:sparse}).\par
    
    The performances of DEEP and the asymmetric version of EP \cite{Scellier2018GeneralizationDynamics} are compared with the same architecture (complete directed graph) and hyperparameters. Asymmetric EP fails to learn XOR and, for AND and OR, we observed that the MSE converged slower than DEEP, which took approximately half the number of epochs, and not always to zero.

\section{Conclusions and Future Work} \label{sec:discussion_conclusions}
    \vspace{-0.1cm}
    In this paper, we extended the \textit{equilibrium propagation} (EP) learning framework towards more bio-plausible artificial neural networks by generalizing its architecture to a complete directed graph and introducing a new neuronal dynamics and learning rule. We also proposed a sparsity-inducing method to prune irrelevant connections. The resulting model was termed \textit{DirEcted Equilibrium Propagation} (DEEP). Simulation results suggest that DEEP is able to learn logical operations that previous models are unable to learn. We supported our results with theoretical sufficient conditions to attain local asymptotic stability during inference.\par 
    
    As DEEP is defined by a continuous-time dynamical system, it provides an interesting line of research for algorithms that can be efficiently implemented with neuromorphic hardware \cite{Ambrogio2018Equivalent-accuracyMemory}. Moreover, due to its unrestricted architecture, DEEP could be used as a network design tool: the optimized, possibly minimalist, structure of the trained networks could be used as an initial architecture for other learning algorithms.\par
    
    To accelerate the convergence of the first phase, a possible idea would be to initialize the neuronal activity as the solution of the neuronal dynamical system obtained when the non-linearity inserted by the hard-sigmoid is removed. Moreover, a detailed theoretical study should be made regarding the approximation scheme used to obtain the results in this discrete framework, in order to determine the influence of the hard-sigmoid in the stability of the equilibrium states reached with respect to this discrete dynamics.\par
    
    It would also be interesting to study whether DEEP can be adapted to a spiking neural network (as in \cite{OConnor2019TrainingPropagation}) and, by leveraging on DEEP's recurrent nature, if it could be used for sequence prediction problems.

\begin{footnotesize}

\end{footnotesize}


\begin{thebibliography}{9}

\bibitem{Rumelhart1987LearningPropagation}
D.E. Rumelhart and J.L. McClelland. Learning internal representations by error propagation. In Parallel Distributed Processing: Explorations in the Microstructure of Cognition: Foundations, chapter 8, pages 318-362. MIT Press, 1st edition, 1987.

\bibitem{Lecun2015DeepLearning}
Y. Lecun, Y. Bengio, and G.E. Hinton. Deep learning. Nature, 521(7553):436–444, 2015.

\bibitem{Guerguiev2017TowardsDendrites}
J. Guerguiev, T. Lillicrap, and B. Richards. Towards deep  learning with segregated dendrites. eLife, 6(e22901), 2017.

\bibitem{Scellier2017EquilibriumBackpropagation}
B. Scellier and Y. Bengio. Equilibrium propagation: bridging the gap between energy-based models and backpropagation. Frontiers Computational Neuroscience, 11(24), 2017.

\bibitem{OReilly1998SixCognition}
R. O’Reilly. Six principles for biologically based computational models of cortical cognition. Trends in Cognitive Sciences, 2(11):455–462, 1998.

\bibitem{Scellier2018GeneralizationDynamics}
B. Scellier, A. Goyal, J. Binas, T. Mesnard, and Y. Bengio. Generalization of equilibrium propagation to vector field dynamics. Arxiv:1808.04873v1, 2018.

\bibitem{Khan2018BidirectionalWaterloo}
A. Khan. Bidirectional learning in recurrent neural networks using equilibrium propagation (master thesis), University of Waterloo, 2018.

\bibitem{Grossberg1987CompetitiveResonance}
S. Grossberg. Competitive learning: from interactive activation to adaptive resonance. Cognitive Science, 11(1):23–63, 1987.

\bibitem{OConnor2019TrainingPropagation}
P. O’Connor, E. Gavves, and M. Welling. Training a spiking neural network with equilibrium propagation. JMLR, 89:1516–1523, 2019.

\bibitem{LeCun2010MNISTDatabase}
Y. LeCun and C. Cortes. MNIST handwritten digit database, 2010.

\bibitem{Ernoult2019UpdatesInput}
M. Ernoult, J. Grollier, D. Querlioz, Y. Bengio, and B. Scellier. Updates of equilibriumprop match gradients of backprop through time in an RNN with static input. In NeurIPS, 2019.

\bibitem{Bartunov2018AssessingArchitectures}
S. Bartunov, A. Santoro, B. Richards, L. Marris, G. Hinton, and T. Lillicrap. Assessing the scalability of biologically-motivated deep learning algorithms and architectures. In NeurIPS, 2018.

\bibitem{Xie1999Spike-basedActivity}
X. Xie and H. Seung. Spike-based learning rules and stabilization of persistent neural activity. In NIPS, 1999.

\bibitem{Hirase2001FiringExperience}
H. Hirase, X. Leinekugel, A. Czurkó, J. Csicsvari, and G. Buzsáki.  Firing rates of hippocampal neurons are preserved during subsequent sleep episodes and modified by novel awake experience. PNAS, 98(16):9386–9390, 2001.

\bibitem{Bejarano2018ACircles}
 D. Bejarano, E.I. Mondragon, and E.G. Hernandez. A stability test for non linear systems of ordinary differential  equations based on the Gershgorin circles. Contemporary Engineering Sciences, 11(91):4541–4548, 2018.

\bibitem{Ambrogio2018Equivalent-accuracyMemory}
S. Ambrogio, P. Narayanan, H. Tsai, R.M. Shelby, I. Boybat,  C. Di Nolfo, S. Sidler, M. Giordano, M. Bodini, N.C.P.  Farinha, B. Killeen, C. Cheng, Y. Jaoudi, and W.G. Burr. Equivalent-accuracy accelerated neural-network training using analogue memory. Nature, 558(7708):60–67, 2018.

\end{thebibliography}
\end{document}